%% file: main.tex
\documentclass[sigconf]{acmart}
\usepackage{hyperref}

\acmConference[GECCO’22]{: Genetic and Evolutionary Computation Conference}{July 9--13}{Boston, MA, USA}

\AtBeginDocument{%
  \providecommand\BibTeX{{%
    \normalfont B\kern-0.5em{\scshape i\kern-0.25em b}\kern-0.8em\TeX}}}

\begin{document}

\title{Empowered Neural Cellular Automata}

\author{Caitlin Grasso}
\affiliation{%
  \institution{University of Vermont}
  \city{Burlington}
  \state{Vermont}
  \country{USA}
}
\email{Caitlin.Grasso@uvm.edu}

\author{Josh Bongard}
\affiliation{%
  \institution{University of Vermont}
  \city{Burlington}
  \state{Vermont}
  \country{USA}}
\email{Josh.Bongard@uvm.edu}

%% ABSTRACT
\input{abstract}

%% CCS CONCEPTS


%%
%% KEYWORDS

\maketitle

% BODY TEXT
\input{introduction}

\input{related_work}

\input{methods}

\input{experimental_design}

\input{results}

\input{discussion}

\input{conclusions}

%% EQUATION EXAMPLES
% \begin{displaymath}
%   \sum_{i=0}^{\infty} x + 1
% \end{displaymath}

% \begin{equation}
%   \sum_{i=0}^{\infty}x_i=\int_{0}^{\pi+2} f
% \end{equation}

%% ACKNOWLEDGEMENTS
\begin{acks}
This material is based upon work supported by the National Science Foundation Graduate Research Fellowship Program under Grant No. 1842491. Any opinions, findings, and conclusions or recommendations expressed in this material are those of the author(s) and do not necessarily reflect the views of the National Science Foundation. We also thank the Vermont Advanced Computing Core for providing the computational resources for this work.
\end{acks}

%% REFERENCES
\bibliographystyle{ACM-Reference-Format}
\bibliography{references}

\end{document}

%% file: abstract.tex
\begin{abstract}
  Information-theoretic fitness functions are becoming increasingly popular to produce generally useful, task-independent behaviors. One such universal function, dubbed empowerment, measures the amount of control an agent exerts on its environment via its sensorimotor system. Specifically, empowerment attempts to maximize the mutual information between an agent's actions and its received sensor states at a later point in time. Traditionally, empowerment has been applied to a conventional sensorimotor apparatus, such as a robot. Here, we expand the approach to a distributed, multi-agent sensorimotor system embodied by a neural cellular automaton (NCA). We show that the addition of empowerment as a secondary objective in the evolution of NCA to perform the task of morphogenesis, growing and maintaining a pre-specified shape, results in higher fitness compared to evolving for morphogenesis alone. Results suggest there may be a synergistic relationship between morphogenesis and empowerment. That is, indirectly selecting for coordination between neighboring cells over the duration of development is beneficial to the developmental process itself. Such a finding may have applications in developmental biology by providing potential mechanisms of communication between cells during growth from a single cell to a multicellular, target morphology. Source code for the experiments in this paper can be found at:  \url{https://github.com/caitlingrasso/empowered-nca}.
\end{abstract}

%% file: introduction.tex
\section{Introduction}

\begin{figure}[!t]
  \centering
  \includegraphics[width=0.5\textwidth]{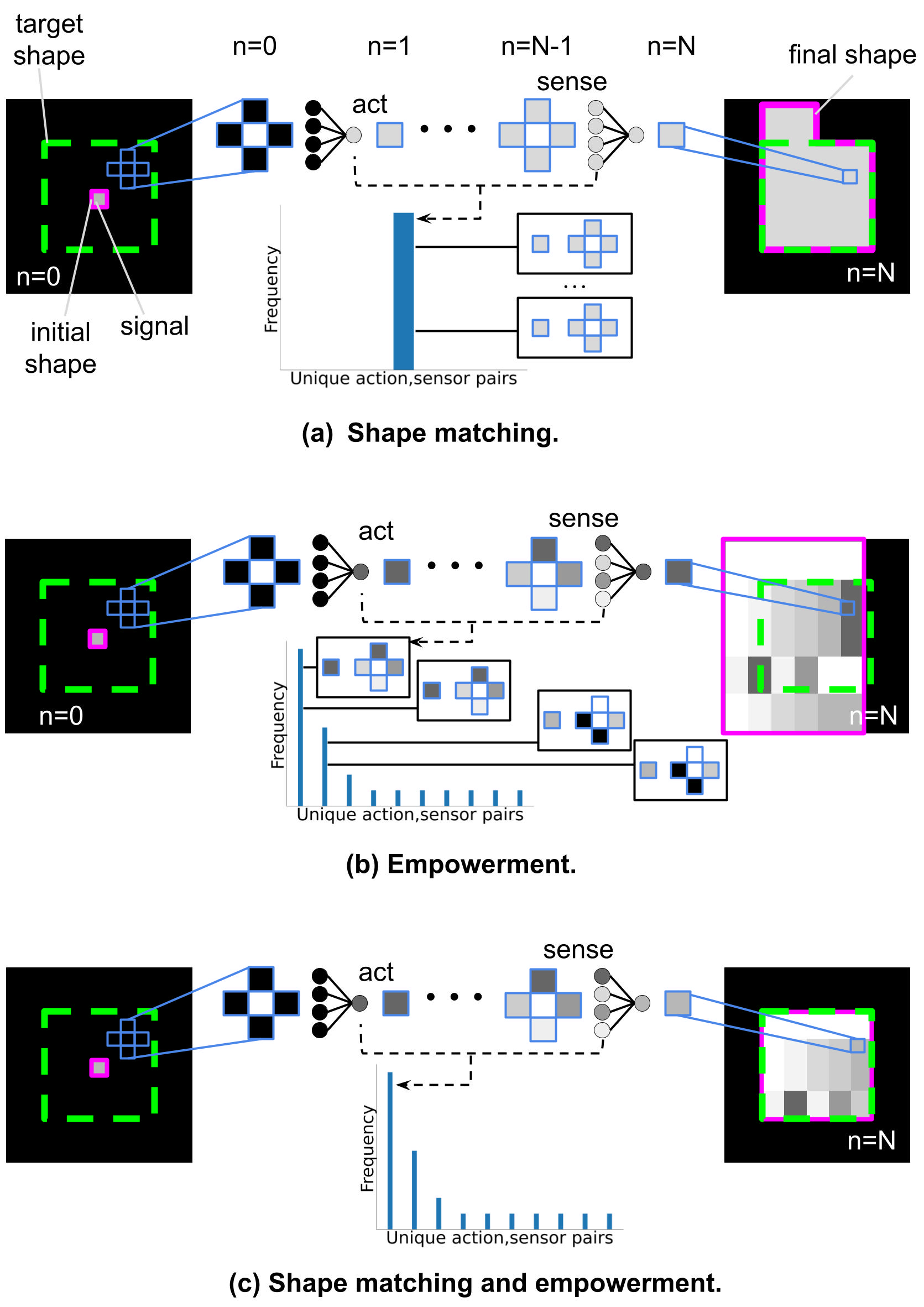}
\caption{(a): Neural cellular automata (NCA) are evolved to grow from an initial seed into a desired target shape (green). The CA's signaling channel (greyscale) can be used to share information between cells but remains constant during simulation. (b) NCA are evolved for empowerment maximize mutual information between cells' past actions (cell's signal) and cells' future sensors (neighboring signals) resulting in a greater diversity of action, sensor pairs and coordination of the CA's signaling throughout space and time. (c) We here report that NCAs evolved for both shape matching and empowerment are better able to match the target shape than those evolved for shape matching alone.}
  \Description{Figure 1. Cartoon overview of the paper. A neural cellular automata is pictured growing from a single cell and attempting to reach a target shape. It is evaluated by error and empowerment and evolved using a genetic algorithm.}
  \label{fig:teaser}
\end{figure}

Biological development, arguably one of the most complex processes in nature, still admits a host of open questions \cite{Levin2018, Pezzulo2015}. Many of these relate to the as yet unknown processes by which cells and tissues coordinate their actions in space and time.
Although many computational models of morphogenesis have been created \cite{dellaert1994toward, bongard2003evolving, stanley2007compositional},
among the simplest are cellular automata (CA) \cite{Basanta2008, eggenberger1997evolving}. In these models, CAs are trained to grow toward and maintain some desired pattern, including deliberate excision of parts of those patterns during growth or after equilibrium as been reached \cite{miller2004evolving}.

Achieving robust morphogenesis in cellular automata requires sophisticated rule sets. This has in part spurred the development of neural cellular automata (NCA) in which rules are replaced by a neural network: the states of a cell's neighbor are fed into a neural network as input, and the cell's new state is supplied at the output \cite{Mordvintsev2020}. The incorporation of deep learning methods into cellular automata also incorporates the need to formulate appropriate loss functions to achieve morphogenesis. To date, loss functions reported in the literature directly reward target pattern fidelity and stability. Such functions are specific to the target shape and and do not generalize to more diverse tasks that may also require coordination of a cell collective. 

In other fields, it has been found that fitness functions that select for agent dynamics seemingly unrelated to the desired behavior often result in the evolution of the desired behavior anyway. Such approaches range from diversity \cite{schmidt2011age, Doncieux2010} and novelty \cite{Lehman2011} metrics to information theoretic measures \cite{Edlund2011, lehman2012rewarding, lehman2013encouraging, Roli2019}. In one such study \cite{For2022}, an information theoretic-based fitness function was employed to increase diversity. Information theoretic fitness functions have proved particularly useful in the domain of evolutionary robotics \cite{bongard2013evolutionary}, where information flows between the agent's sensory system, motor system and environment often result in coordinated, interesting and/or useful robot behaviors \cite{Der2008, Ay2008, Eysenbach2019}.

This interest in information flows within robots mirrors an interest in information flows within organisms. Neuroscience has a long history of using information theory to study functional connectivity in the brain \cite{bullmore2009complex} and abstract cognitive phenomena \cite{tononi2016integrated}, but there is also some work on using information theory to study the more grounded aspects of intelligent behavior: sensorimotor coordination \cite{mulliken2008forward, kokal2010granger}. Robots and cellular automata are simplified models of biological organisms and processes respectively, and are thus ideal substrates within which to study growth and behavior from an information theoretic point of view.

One particular information theoretic metric of interest for evolutionary robotics and cellular automata alike is that of empowerment: maximizing mutual information between actions in the past and sensor states in the future \cite{Klyubin2005b}. Empowerment is of interest because it describes, in information theoretic terms, a form of control and coordination that can extend over space and time: an agent is empowered if it can act in such a way as to influence particular outcomes among its near or possibly distant neighbors, over increasingly long time horizons. Given this, we hypothesize that selecting for empowerment within an evolving population of CAs may guide them toward the desired behavior of matching a target shape, a simplification of biological morphogenesis, as morphogenesis likely requires large-scale and long-term control and/or coordination. 

There are many ways one might attempt to empower a CA. Here, we start with the most obvious approach: we treat each cell in the CA as an agent, and evolve for maximimal mutual information between cells' actions early in development and cells' sensor states later during during development. As described in the methods section below, we apply a multiobjective optimization approach which allows us to evolve for diversity within the CA population, morphogenesis, empowerment, or any subset of these three. 

As we report in the result section, we find here that CAs evolved for morphogenesis and empowerment exhibit better morphogenesis than CAs evolved for morphogenesis alone. This supports the findings from the evolutionary robotics literature that evolving for empowerment can indirectly exert selection pressure for desired behaviors that require sensor/motor coordination. In the discussion section we investigate the patterns of growth, signaling and empowerment in several random and evolved CAs: the lack of any obvious relationships there indicate that there remain many subtle interdependencies between information flows within growing cellular automata and developing organisms yet to be investigated.

%% file: related_work.tex
\section{Related Work}

This work builds off of seminal work on empowerment by applying the concept to a cellular automaton system. Empowerment was first introduced by Klyubin et al. \cite{Klyubin2005a, Klyubin2005b} as a local, universal, task-independent intrinsic motivation that can be used as the sole fitness function in the evolution of an agent's sensor/actuator placements to achieve a task. For an in-depth review of empowerment, we refer the reader to \cite{Salge2014}. 

Most empowerment work to date focuses on an individual agent, however, of particular interest are the few studies that consider a multi-agent scenario. Capdepuy et al. \cite{Capdepuy2007, Capdepuy2012} investigated the interactions between agents that individually attempt to maximize empowerment in a shared environment and found that interesting structures can emerge. Similarly, Guckelsberger and Polani \cite{Guckelsberger2014} evolved agents for empowerment maximization in a resource-centric environment which resulted in agents exhibiting biologically plausible behaviors, such as greed and parsimony. Lastly, Clements and Polani \cite{Clements2017} introduce the idea of team empowerment. They showed that maximizing empowerment of a sports team of agents in an Ultimate frisbee simulation can produce recognizable team sport behaviors. In all current work on multi-agent empowerment, agents are considered in the traditional robotic sense with a conventional sensorimotor apparatus. Additionally, with the exception of team empowerment, agents work to maximize their individual empowerment as opposed to that of a collective. We expand the approach to a non-traditional sensorimotor apparatus embodied by a neural cellular automata where each cell is an agent with sensors/actuators defined by the input/output of the neural network rules. Cells in the CA not only share the same environment but also are working together to achieve a global task.

The NCA model presented here is largely inspired by two cellular automaton models of morphogenesis. First, Miller \cite{miller2004evolving} used a boolean feed-forward circuit to represent the rules of a cellular automata and evolved these circuits to produce CA capable of growing from a single cell into a French flag. The second is Mordvintsev et al. \cite{Mordvintsev2020} who introduced neural cellular automata and successfully trained a CNN to learn the update rules for a CA capable of growing from a single cell into complex shapes.

Recently, there have been many works that use NCA models to achieve morphogenesis in various platforms. Most rely on objective functions that aim to minimize the distance between the current shape of the cellular automaton and a pre-specified target shape \cite{Nichele2018, Sudhakaran2021, Zhang2021, Moore2018, Horibe2021, chan2018lenia}. Others are purely open-ended with no particular target shape, aiming only to generate interesting structures and/or behaviors \cite{Wang}. Here, we employ a simple neural network to achieve the task of shape matching. Rather than adding complexity to the network to better achieve the task, we introduce empowerment as an additional objective and hypothesize that NCA intrinsically motivated by empowerment are better capable of morphogenesis than those that are not.

%% file: methods.tex
\section{methods}

We evolve a population of NCAs that attempt to grow from a single cell into a target shape, a simplified version of biological morphogenesis. NCAs are evaluated based on their ability to match the target shape and/or their empowerment during development. Those NCAs scoring poorly on the specified objective(s) are removed from the population while those scoring well are randomly modified to produce the next generation. Evolution thus produces NCAs that are capable of shape matching, highly empowered, or both.   

\begin{figure}[!t]
  \centering
  \includegraphics[width=\linewidth]{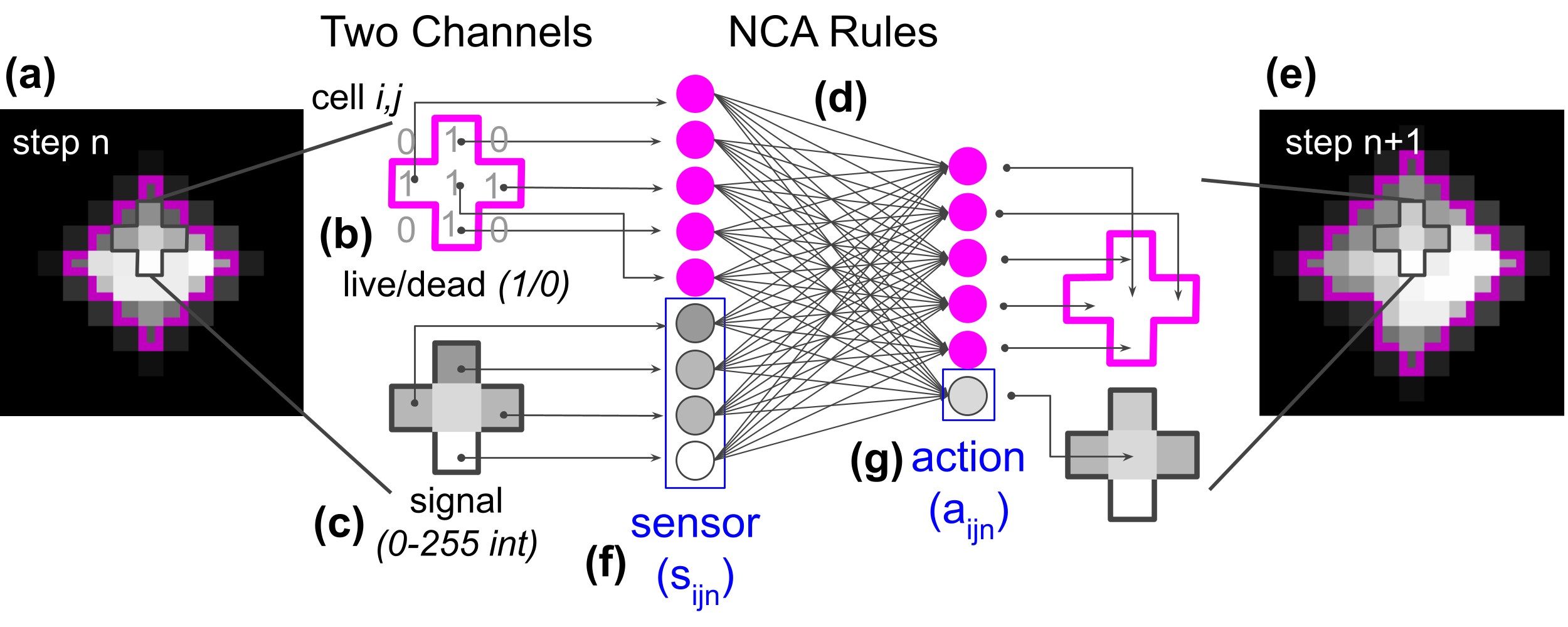}
\caption{A single update of the NCA model. The NCA at an arbitrary time step $n$ (a) is comprised of two channels: a binary live/dead channel (b) and a signaling channel (c). At each time step, a neural network (d) is applied sequentially to each cell in the CA. For cell $i,j$, the network takes as input the neighboring cells' live/dead values and signals. The output of the network produces the updated the CA grid at time step $n+1$ (e). In computing empowerment, each cell is considered an agent. The sensor state of cell $i,j$ at time step $n$, $s_{ijn}$, is the average signal of the cell's neighbors (f). Cell $i,j$ produces an action at time step $n$, $a_{ijn}$, which updates its own signal value for neighboring cells to sense (g).}
  \Description{Figure 2. Describes the NCA model and is fully described in the text.}
  \label{fig:methods}
\end{figure}

\subsection{NCA Model}

The NCA model is a discrete cellular automaton consisting of two channels (Figure \ref{fig:methods}a): a binary live/dead channel (Figure \ref{fig:methods}b), and a signaling channel of integers between 0-255 (Figure \ref{fig:methods}c). Similar to signaling mechanisms in biological organisms, this second channel can be used by cells to share additional information with one another. 

The rules of the CA are embodied by a feed-forward neural network with no hidden layers (Figure \ref{fig:methods}d). The inputs for a given cell include neighboring cells' signals and the live/dead values for neighboring cells and the cell itself. The network outputs an updated live/dead value and signal for the cell itself and four binary values that determine whether the cell should replicate it's states into each of it's four neighbors. These binary replication outputs allow for growth or death of contiguous patches of cells. Outputs of the network pass through a sigmoid activation function and are then binarized or scaled depending on the output type. For the binary outputs, if the activation is greater than $\epsilon$, the output is $1$, otherwise it is set to $0$. For the signal concentration node, if the output is greater than $\epsilon$, the value is scaled between
$[127,255]$ and binned to an integer, otherwise it is set to $0$. In this way, the NCA is heavily biased towards growth and signaling. As in biological morphogenesis, proliferation and signaling are more common than cell death or quiescence making controlled growth a challenging task.

At each time step, the neural network is executed sequentially for each cell in the grid to determine the global state of the CA at the following time step (Figure \ref{fig:methods}e). In a similar manner to biological signaling gradients, signal diffuses through the CA at each time step at a decaying rate. Each CA simulation begins with a seed cell at the center of an $M \times M$ grid with a signal value of zero. The CA develops for $N$ time steps at which point the resulting morphology can be compared to a pre-determined target shape. 

\subsection{NCA Evaluation}

It is currently unknown what cellular communication strategies are required to ensure faithful morphogenesis in nature, but it is likely that large-scale coordination is necessary. This concept of coordination can be efficiently captured by the information theoretic concept of empowerment \cite{Klyubin2005b}: the actions of an agent in the present yields high mutual information with the states of other agents at a future time. To that end, two functions are used to evaluate NCAs: loss, which evaluates the CA's morphogenetic capability by computing the distance between a target shape and the final shape of the CA after $N$ time steps, and empowerment, computed from a CA's signaling dynamics during development. 

\subsubsection{Loss} To determine how well the CA matches a static, pre-defined target shape, $T$, the L2 loss, $\mathfrak{L}$, is computed between the binary state of the $M$-dimensional CA at a given time step, denoted by $C_n$, and the target averaged over the number of CA time steps starting at time step $n_0$ and terminating at step $n_1$. Loss, as defined in Equation \ref{eqn:error}, is to be minimized. 

\begin{equation}
    \mathfrak{L}(n_0, n_1) = \frac{1}{n_1-n_0} \sum_{n \in [n_0, n_1]} \frac{\sum_{i,j \in M} (C_{ijn}-T_{ij})^2}{M^2}
    \label{eqn:error}
\end{equation}

\subsubsection{Empowerment}

We expand empowerment, as defined in \cite{Klyubin2005b}, to NCA by defining an agent as a single cell. The environment of the cell is defined by the rest of the CA grid. Cells' only have direct influence over their neighbors via CA neural network ruleset. Empowerment is applied only to the signaling dynamics of the CA. We define the sensor state of cell $i,j$ at time step $n$, $s_{ijn}$, as the average signal of its Von Neumann neighbors (Figure \ref{fig:methods}f). The action state of cell $i,j$ at time step $n$, $a_{ijn}$, is the cell's own signal value as output from the neural network (Figure \ref{fig:methods}g). 

Parallel to a traditional robotic sensorimotor system, the cell can be thought of as an agent with four sensors (one on each side) facing outward to sense the signals of its neighbors and one actuator which acts on its environment by updating the signal the cell itself emits for neighboring cells to sense. We define empowerment as maximizing the mutual information between the set of all cells' action states during one time period in CA development and the corresponding set of all cells' sensor states during an equal-duration time period later in development. These time periods were arbitrarily chosen to be the first $N/2$ time steps and the last $N/2$ time steps where $N$ is the total number of developmental time steps. Action ($A$) and sensor ($S$) sets are defined as follows where $M$ is the dimension of the CA grid. 

\begin{displaymath}
  A_0^{N/2} = \{ a_{000},\cdots, a_{MM\frac{N}{2}}\}\\
\end{displaymath}

\begin{displaymath}
    S_{N/2}^{N} = \{ s_{00\frac{N}{2}},\cdots, a_{MMN}\}
\end{displaymath}

Thus, empowerment, $\mathfrak{E}$, measured in bits, is the mutual information between these sets of actions and sensor states and can be maximized via evolutionary search.

\begin{equation}
    \mathfrak{E} = I(A_0^{N/2}, S_{N/2}^{N}) = -\sum_{a_{ijn},s_{ijn}} p(a_{ijn},s_{ijn}) \log_2 \frac{p(a_{ijn},s_{ijn}) }{p(a_{ijn})p(s_{ijn})}
    \label{eqn:empowerment}
\end{equation}

% Mutual information between the CA's action set and sensor set is defined as 

% \begin{displaymath}
%     I(A_0^{N/2}, S_{N/2}^{N}) = -\sum_{a_{ijn},s_{ijn}} p(a_{ijn},s_{ijn}) \log_2 \frac{p(a_{ijn},s_{ijn}) }{p(a_{ijn})p(s_{ijn})}
% \end{displaymath}

% From an information theoretic perspective, channel capacity is the maximum data transmission rate across a channel and is computed as the maximum mutual information between the input and output data. Thus, empowerment, $\mathfrak{E}$, of the NCA is by definition the maximum mutual information between the CA's set of actions and corresponding set of sensor states later in development.

% \begin{equation}
%     \mathfrak{E} = I(A_0^{N/2}, S_{N/2}^{N})
%     \label{eqn:empowerment}
% \end{equation}

Empowerment, as stated in Equation \ref{eqn:empowerment}, was implemented using PyInform, a Python package for information-theoretic computation \cite{Moore2018_inform}. Empowerment is to be maximized, however, for ease of implementation it was converted into a minimization function by multiplying by negative one. In the following figures, empowerment is displayed as a maximization function by undoing this negation. 

% LOCAL MUTUAL INFORMATION ???

% GLOBAL VS. LOCAL EMPOWERMENT ???

% We also attempted a local version of CA empowerment by computing mutual information between a single cell's action states in the first half of simulation and sensor states in the second half of simulation and the averaging over all cells to determine empowerment of the entire CA simulation, however, this method proved to be less beneficial than its global counterpart. 

%% file: experimental_design.tex
\section{Experimental Design}

NCAs were evolved using Age-Fitness Pareto Optimization (AFPO) \cite{schmidt2011age}, a multi-objective evolutionary algorithm which uses age of genetic material as an additional objective to maintain diversity in a population. Four variations of AFPO outlined in Table \ref{tab:treatments} were tested with either one or two additional objectives (age is always the first objective). To rule out the possibility that adding more objectives makes evolutionary search for this task easier, we include an tri-objective loss-only control ensuring a fair comparison with the tri-objective loss-empowerment treatment. This is achieved by splitting loss (Equation \ref{eqn:error}) into the first half of CA development and the second half of CA development.

\begin{table}[h]
    \centering
    \begin{tabular}{|c|c|c|c|}
    \hline
     \textbf{Treatment} &  \textbf{Obj. 1} & \textbf{Obj. 2} & \textbf{Obj. 3}\\
     \hline
     Bi-error & Age & $\mathfrak{L}(0,N)$ [\ref{eqn:error}]& - \\
     \hline
     Tri-error- & Age & $\mathfrak{L}(0,N)$ [\ref{eqn:error}] & $\mathfrak{E}$ [\ref{eqn:empowerment}] \\
     empowerment & & & \\
     \hline
     Tri-error & Age & $\mathfrak{L}(0,N/2)$ [\ref{eqn:error}] & $\mathfrak{L}(N/2,N)$ [\ref{eqn:error}] \\
     \hline
     Bi-empowerment & Age & $\mathfrak{E}$ [\ref{eqn:empowerment}] & - \\
     \hline
    \end{tabular}
    \caption{Objective functions for four different treatments of Age-Fitness Pareto Optimization (AFPO).}
    \label{tab:treatments}
\end{table}

For each treatment, 25 evolutionary runs were performed for 2,000 generations each and with a population size of 400 NCAs. A CA grid dimension of $M=25$ is used and the target shape for all runs is a square centered around the initial seed. During evaluation, the NCA is run for $N=50$ time steps and evaluated on the objectives described in Table \ref{tab:treatments}. Wall time for a single evolutionary run was about 10 hours. Experiments were performed in parallel using CPU resources provided by the Vermont Advanced Computing Core.

%% file: results.tex
\section{Results}

\begin{figure}[!t]
  \centering
  \includegraphics[width=\linewidth]{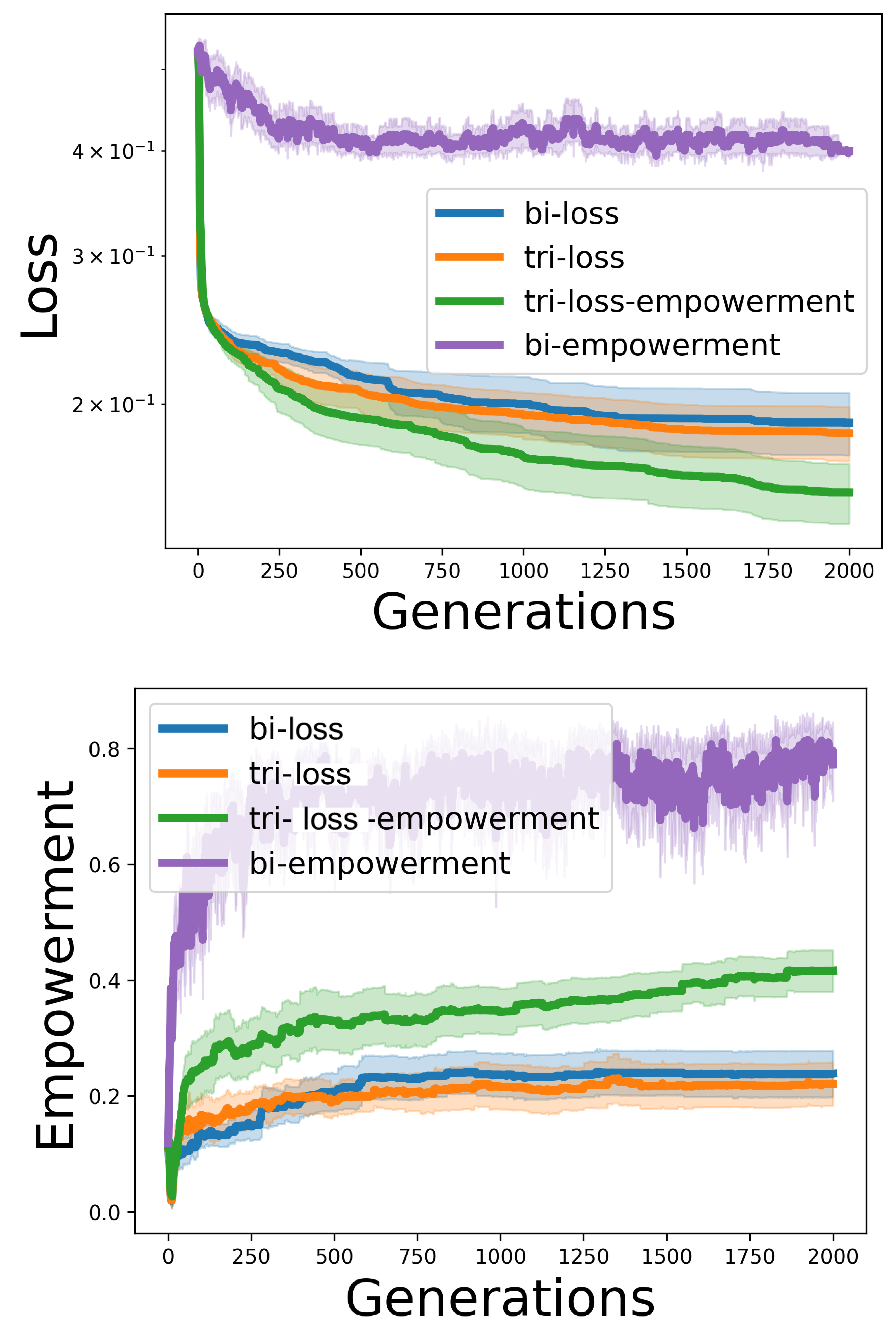}
  \caption{(top): Loss of the lowest loss NCA in the population at each generation averaged over all 25 runs (95\% C.I.) for each AFPO treatment. (bottom): Empowerment of the lowest loss NCA in the population at each generation averaged over all 25 runs (95\% C.I.) for each AFPO variation. Empowerment is measured in bits.}
  \Description{Top plot: Line plot of generations versus loss with four decreasing curves for each AFPO variation. The tri-loss-empowerment curve has the lowest loss, followed by the two loss-only controls, and then the empowerment only runs which have significantly higher loss. Bottom plot:Line plot of generations versus empowerment with four increasing curves for each AFPO variation. The bi-empowerment curve has the highest empowerment, followed by the tri-loss-empowerment lines, and the two loss-only controls have the lowest empowerment.}
  \label{fig:curves}
\end{figure}

\begin{figure}[!t]
  \centering
  \includegraphics[width=\linewidth]{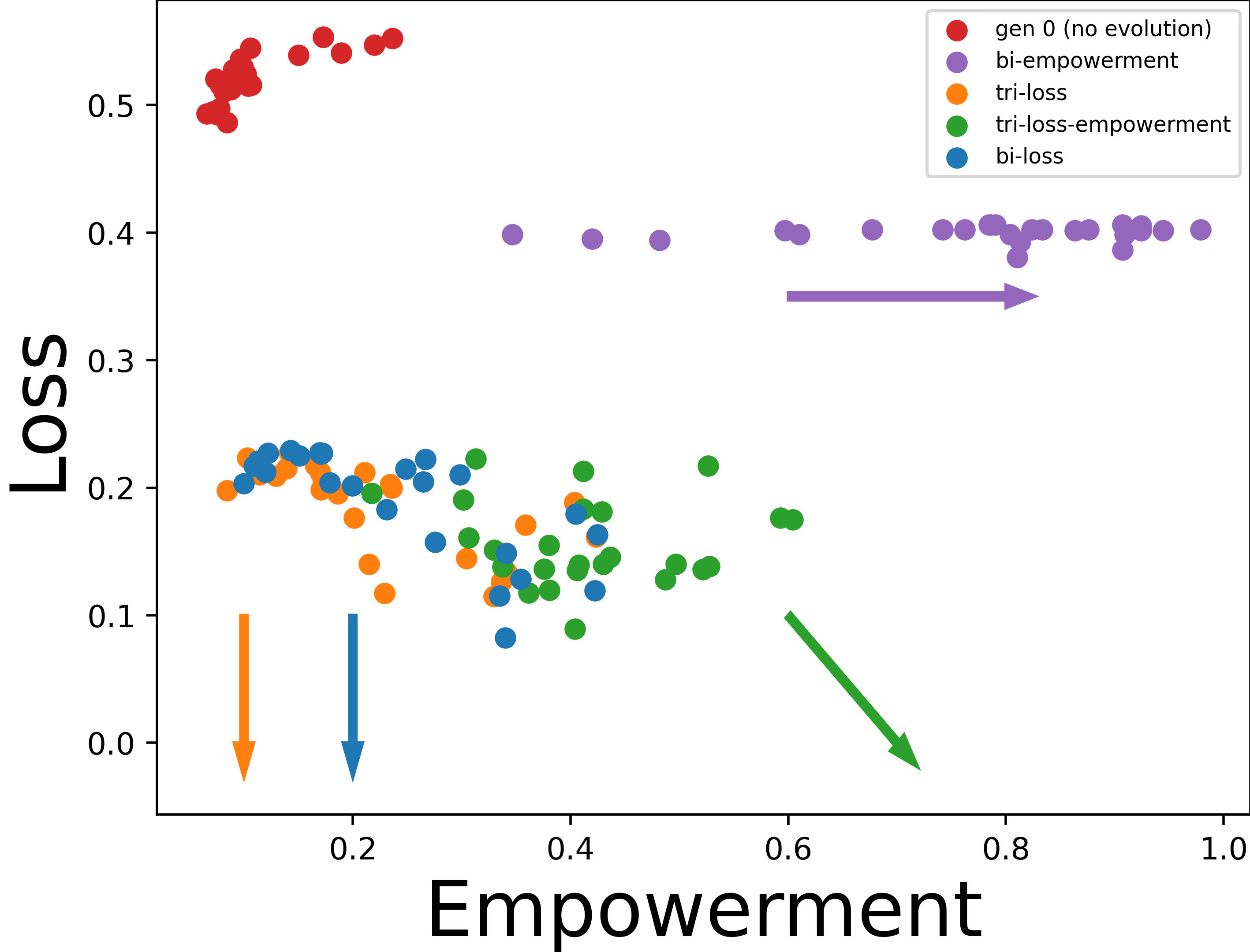}
  \caption{The most lowest loss NCAs resulting from each of the 25 evolutionary runs for each AFPO variation in the loss-empowerment space. Arrows corresponding to each treatment indicate the direction of selection pressure for that treatment. Shown in the top-left in red are the lowest loss NCAs from 25 different populations at generation zero. No evolution occurs on these individuals, thus, there is no corresponding arrow.}
  \Description{Scatter plot with axes empowerment versus loss showing 25 dots per treatment, plus 25 dots for the random NCAs. Arrows point straight down in the negative loss direction for the loss-only controls. The arrow for the tri-loss-empowerment runs points to the bottom left of the plot. Lastly, the arrow for the bi-empowerment runs points to the positive empowerment direction.}
  \label{fig:scatter}
\end{figure}

Results of the evolutionary runs for each treatment are shown in Figure \ref{fig:curves}. The tri-loss-empowerment treatment achieves the lowest loss. That is, NCAs evolved for both loss minimization and empowerment maximization on average come closer to matching the target shape than those evolved for matching the target alone. The bi-loss and tri-loss curves achieve similar performance indicating that the tri-loss treatment is an adequate tri-objective implementation of the bi-loss treatment. Additionally, empowerment of the tri-loss-empowerment runs is higher than that of the loss-only controls suggesting evolution is occurring on the empowerment objective. As expected, the bi-empowerment treatment achieved the highest loss as well as the highest empowerment as NCAs were only trained to maximize empowerment and there was no direct selection pressure for them to grow the target shape. 

 Interestingly, empowerment of the two loss-only treatments (blue and orange curves in Figure \ref{fig:curves} bottom panel), though not directly selected for, increases during evolution. Likewise, the loss of the bi-empowerment treatment (purple curve in Figure \ref{fig:curves} top panel) decreases at the beginning of evolution before flattening. These trends suggest that the tasks of shape matching and empowerment have a synergistic relationship. In other words, evolving for morphogenesis alone naturally produces NCAs with a higher empowerment and vice versa. Thus, introducing direct selection pressure for empowerment pushes the NCAs into part of the fitness landscape that is also beneficial for the shape matching task and evolution speeds up. 
 
The best (lowest loss) NCAs resulting from each of the 25 runs for every treatment were visualized in the loss-empowerment space and compared to 25 random NCAs, shown in Figure \ref{fig:scatter}. Random NCAs were produced by selecting the lowest-loss NCAs in 25 generation zero populations (of size 400). Arrows indicate direction of selection pressure for each treatment. Random NCAs are clustered in the top right and exhibit the highest loss and lowest empowerment. Clusters for the other treatments are positioned in the direction of their selection pressures relative to the random cluster. Notably, loss of the bi-empowerment cluster is lower than that of the random cluster. Also interesting to note is that the loss-only controls are not as clustered near low empowerment compared to random. Again, suggesting that the two objectives are not independent.

%% file: discussion.tex
\begin{figure}[!t]
  \centering
  \includegraphics[width=\linewidth]{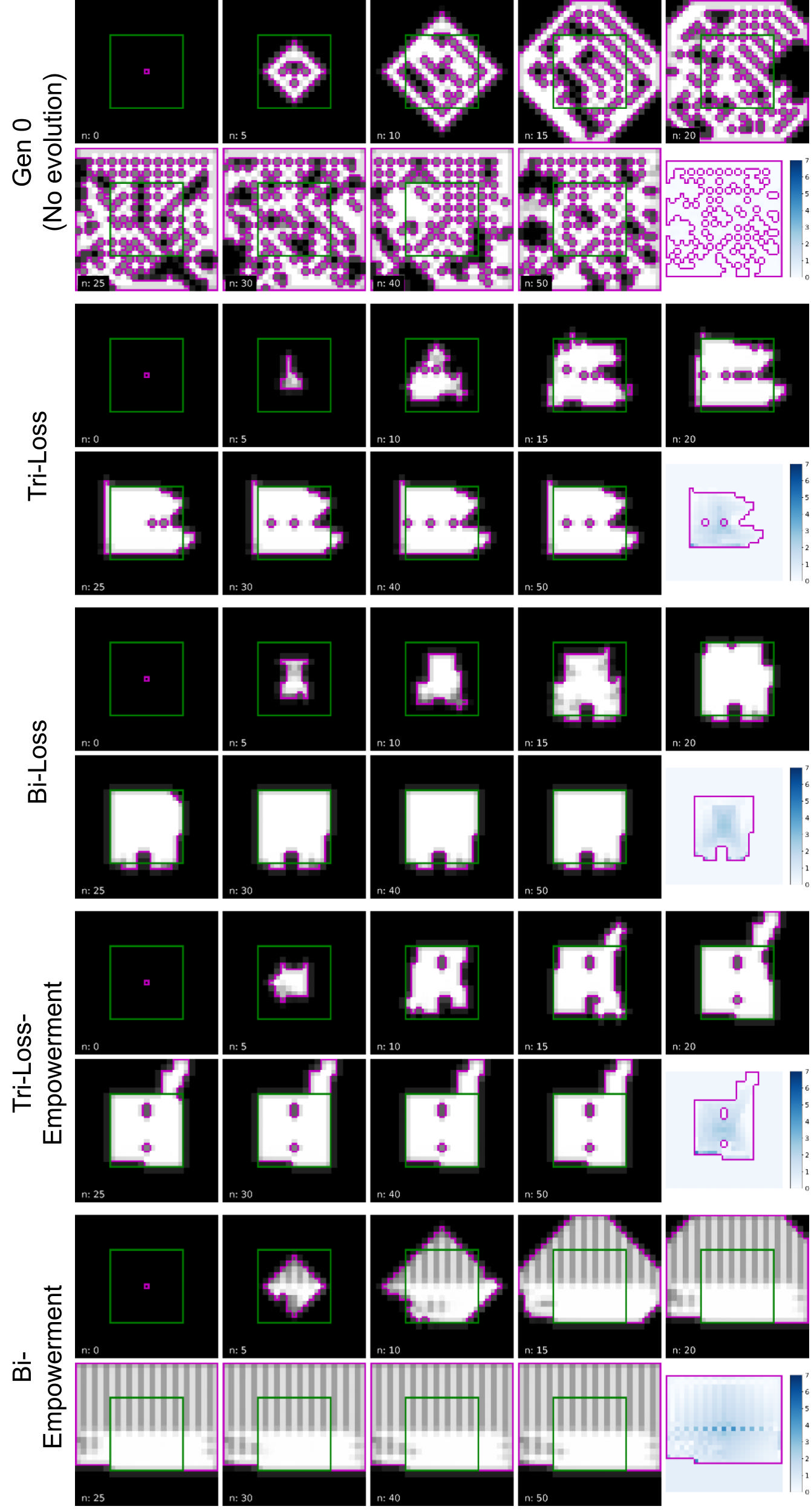}
    \caption{Morphogenesis of the lowest loss NCA over all runs from each treatment starting from the initial seed at $n=0$ to the end of simulation, $n=50$ as it attempts to match the target shape (green). Local mutual information corresponding to each cell's contribution to overall empowerment is depicted with the blue heatmap. Darker blue indications higher mutual information.}
  \Description{Figure 5. Fully described in the text.}
  \label{fig:best_error_film}
\end{figure}

\begin{figure}[!t]
  \centering
  \includegraphics[width=\linewidth]{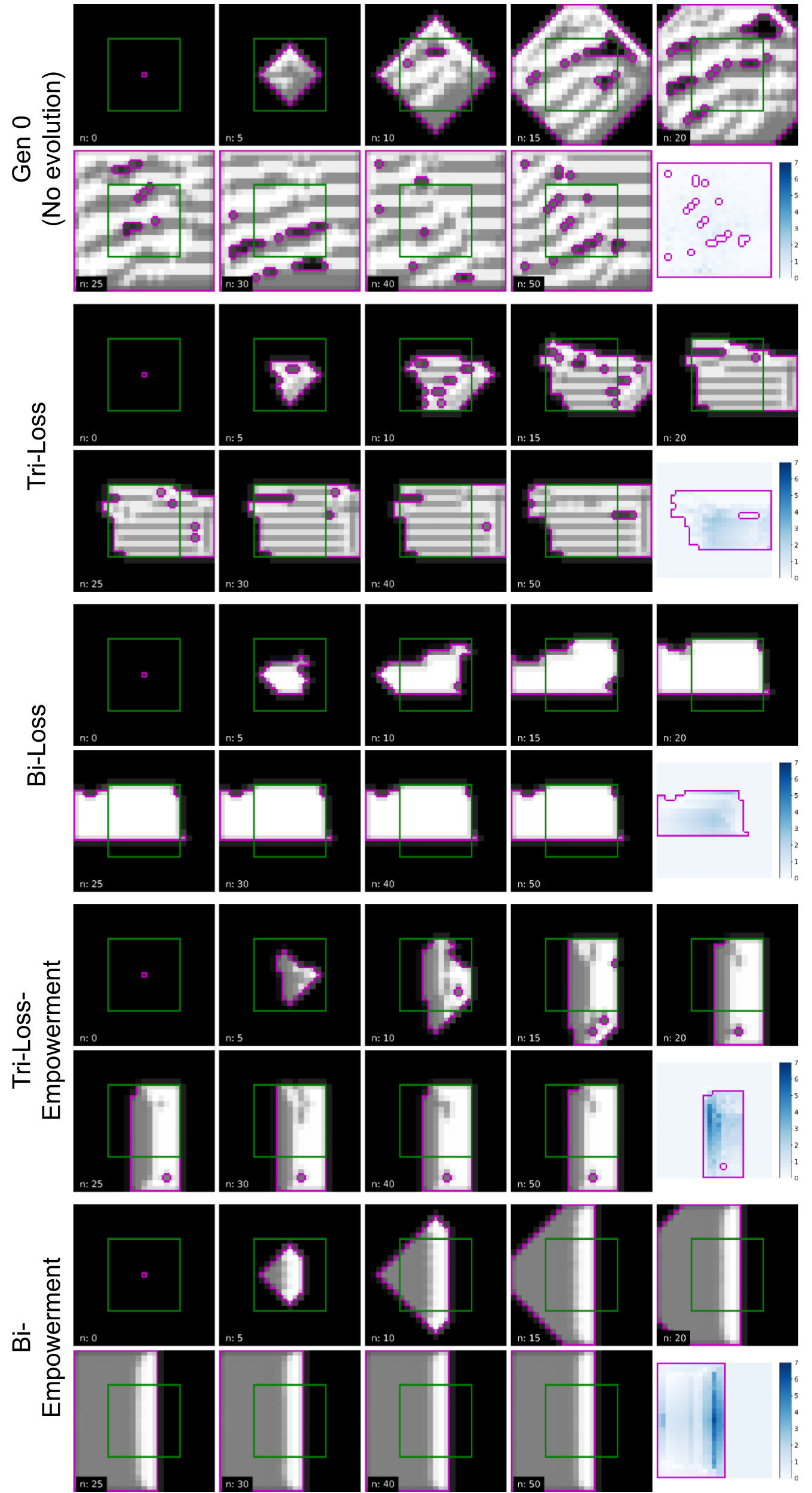}
  \caption{Morphogenesis of the most empowered NCA over all runs from each treatment starting from the initial seed at $n=0$ to the end of simulation, $n=50$ as it attempts to match the target shape (green). Local mutual information corresponding to each cell's contribution to overall empowerment is depicted with the blue heatmap. Darker blue indications higher mutual information.}
  \Description{Figure 6. Fully described in the text.}
  \label{fig:best_empowerment_film}
\end{figure}

\section{Discussion}

The above results provide evidence that adding empowerment as an objective in evolutionary search produces NCAs that are better able to perform the task of morphogenesis than evolving for morphogenesis alone. That is, pushing NCA towards a signaling mechanism that maximizes the amount of information shared between the actions of cells' in the past and the senses of cells at a future time point is useful for the process of morphogenesis. Understanding this signaling mechanism that results from maximizing empowerment is a difficult task due to the number of cells in the CA and the long time horizon between a cell's action and its corresponding sensor value. However, visualizing signaling dynamics over time and local empowerment heatmaps may provide some clues regarding its characteristics. Local empowerment measures the contribution of a single cell's sensor, action pairs to the overall empowerment score of the CA. 

Figure \ref{fig:best_error_film} depicts the development of the lowest-loss NCAs from each treatment. In accordance with the loss curves in Figure \ref{fig:curves}, the bi-loss, tri-loss, and tri-loss-empowerment treatments are best at matching the target shape, though none achieve the task perfectly. These treatments also appear to show similar signaling dynamics with the majority of cells having high signal concentrations. They also yield similar local mutual information heatmaps: corner or edge cells have the highest local mutual information and also seemingly high mutual information in the center of the shape that fades outwards. In contrast, the bi-empowerment NCA appears to make use of a broader range of signaling values which produce a striped pattern. Whether these patterns are indicative of some underlying useful dynamic or are simply artefacts of the CAs chosen for representation remains to be determined.

Striped signaling patterns and those that use a wider range of signal concentrations appear to be more common in NCAs with higher empowerment (Figure \ref{fig:best_empowerment_film}). For the tri-loss-empowerment and bi-empowerment treatments, local empowerment appears highest in cells that appear earlier in development. Growth then proceeds outwards from these regions. Notably, in both Figures \ref{fig:best_error_film} and \ref{fig:best_empowerment_film}, evolved NCAs exhibit more cohesive shapes with fewer holes and more controlled growth than the random NCAs. This is particularly interesting for the bi-empowerment NCAs where shape of the NCA is not considered in the evaluation or selection of NCAs during evolution. 

\begin{figure}
  \centering
  \includegraphics[width=0.8\linewidth]{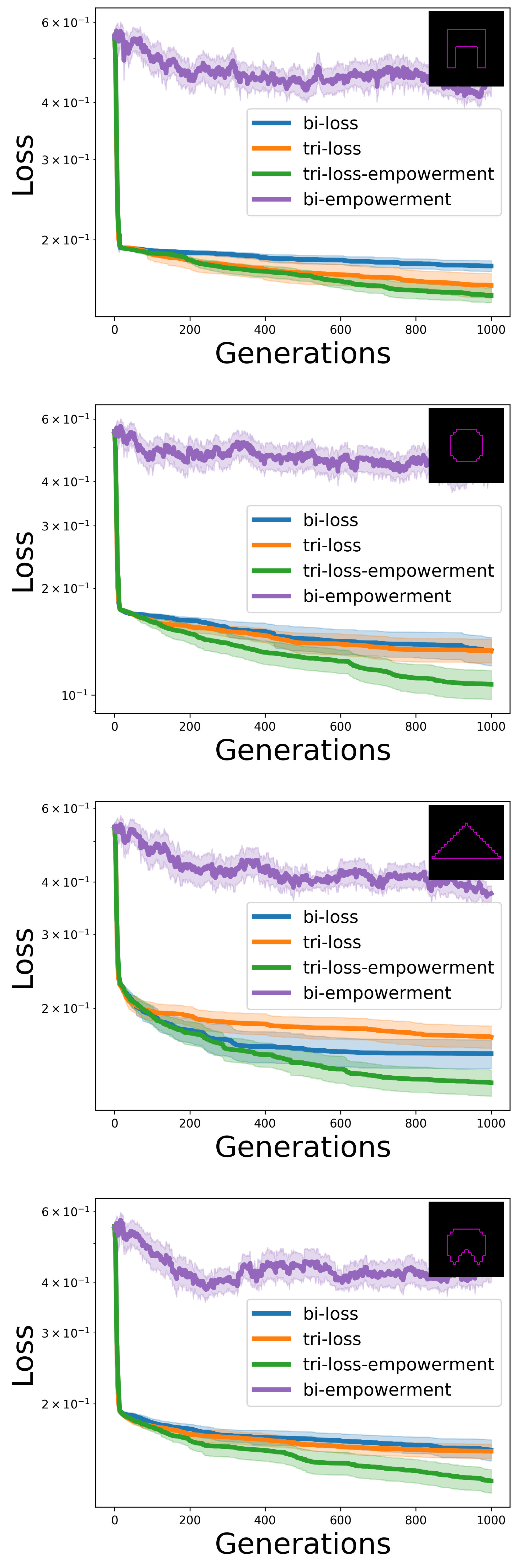}
  \caption{Loss over evolutionary time for four different shapes: biped (top), circle (middle-top), triangle (middle-bottom), and circular biped (bottom).}
  \Description{Figure 7. Four line plots depicting error versus empowerment each corresponding to a different target shape. Each line plot has four lines corresponding to the four treatments as in Fig. 5. All line plots show similar characteristics to those in Figure 5.}
  \label{fig:other_shapes}
\end{figure}

\begin{figure}
  \centering
  \includegraphics[width=\linewidth]{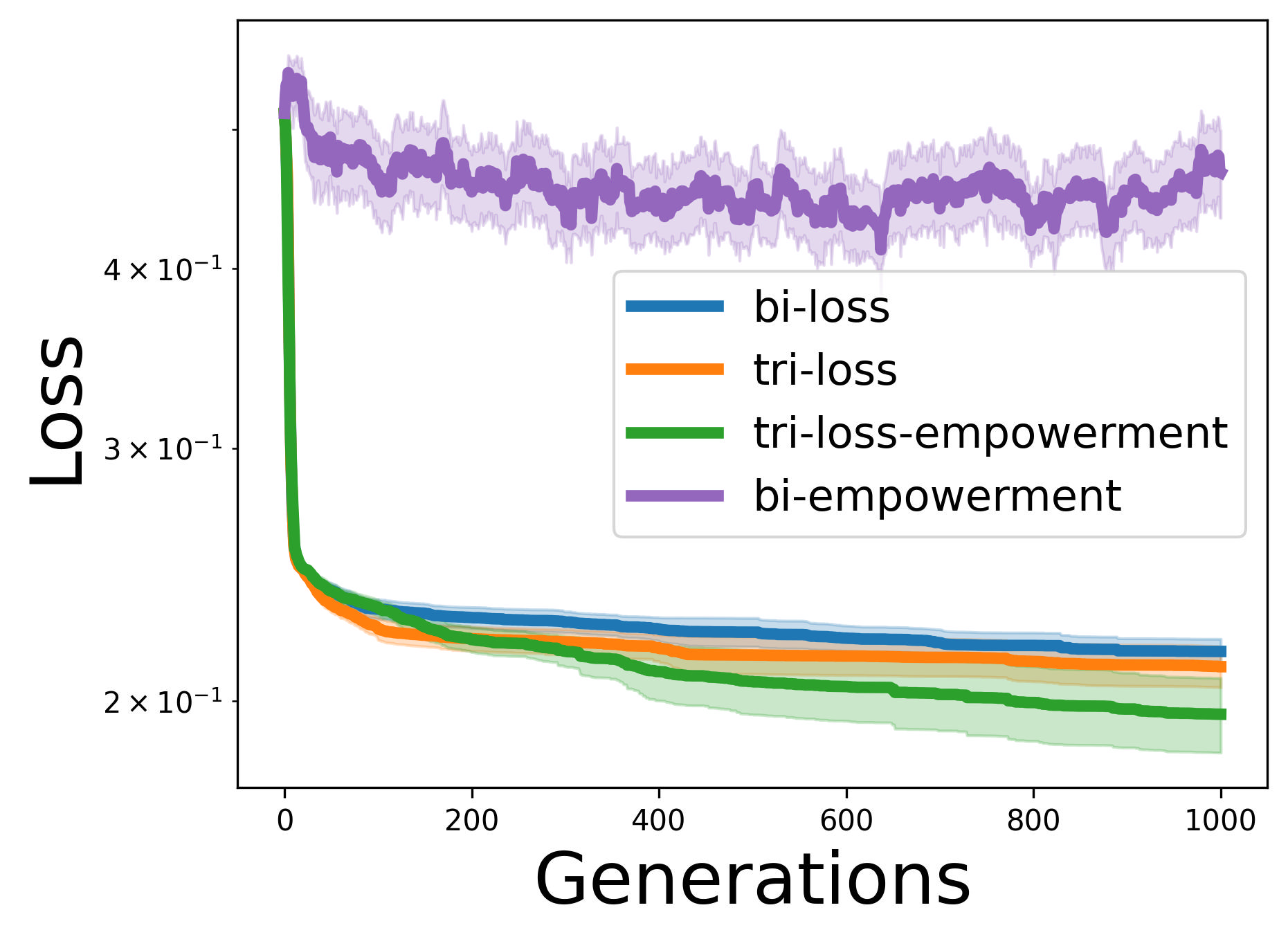}
  \caption{Loss over evolutionary time for a CA with grid resolution 50x50 (double that used in Figure \ref{fig:curves}) and a square target shape.}
  \Description{Figure 8. Single plot depicting error versus empowerment curves for a CA with double the grid resolution and a square target. The plot has four lines corresponding to the four treatments as in Fig. 3. All line plots show similar characteristics to those in Figure 3.}
  \label{fig:high_res}
\end{figure}

To explore whether results are specific to the square target shape and $25 \times 25$ grid resolution, experiments were repeated for four different target shapes (triangle, circle, biped, and circular biped) and at double the resolution ($50\times50$ grid). Figure \ref{fig:other_shapes} displays loss curves for four different target shapes while Figure \ref{fig:high_res} displays those for the square target on a grid of double the original resolution. Experiments were conducted as described in Section 3 with the exception of the number of generations which was decreased from 2,000 to 1,000 for computational reasons. Loss curves in all scenarios display similar trends to those in Figure \ref{fig:curves} suggesting that empowerment is beneficial for and generalizes to various morphogenetic processes and increasingly complex tasks.

%% file: conclusions.tex
\section{Conclusions and Future Work}

In this paper, we have expanded the application of empowerment, a universal, information-theoretic objective function, to a neural cellular automaton system whereby a cell collective attempts to coordinate its behavior to achieve a global task. We show that including empowerment as an objective in the evolutionary search of NCAs with the primary task of shape matching results in more performant NCAs compared to selecting for shape matching alone. When cells are selected to exert control over their environment (neighboring cells) via their signaling dynamics they are better able to coordinate their actions through space and time to achieve a collective task. 

This work suggests that there may be certain information-theoretic signatures, such as empowerment, of CA dynamics that produce generally useful, task-independent behaviors. Such a notion has clear parallels to biology in which complex networks of cells share information by means of chemical, bioelectric, and mechanical signals in order to achieve specific tasks, such as morphogenesis. Uncovering these information signatures by means of simulation suggest potential mechanisms by which biological cells communicate and coordinate behavior. Much future work remains to better understand the relationship between empowerment and NCA signaling dynamics including decomposing highly-empowered NCAs and examining their dynamics.
Additionally, it will be interesting to investigate different variations of empowerment (altering the time horizon, expanding the cells' neighborhood, etc.) as well as examine the impact of empowerment on evolutionary search for more diverse CA tasks. 